# A Real-time Spatio-Temporal Trajectory Planner for Autonomous Vehicles with Semantic Graph Optimization

Shan He[1,2], Yalong Ma[1,2], Tao Song[1,2], Yongzhi Jiang[1,2] and Xinkai Wu[1*]

*Abstract*—Planning a safe and feasible trajectory for autonomous vehicles in real-time by fully utilizing perceptual information in complex urban environments is challenging. In this paper, we propose a spatio-temporal trajectory planning method based on graph optimization. It efficiently extracts the multi-modal information of the perception module by constructing a semantic spatio-temporal map through separation processing of static and dynamic obstacles, and then quickly generates feasible trajectories via sparse graph optimization based on a semantic spatio-temporal hypergraph. Extensive experiments have proven that the proposed method can effectively handle complex urban public road scenarios and perform in real time. We will also release our codes to accommodate benchmarking for the research community

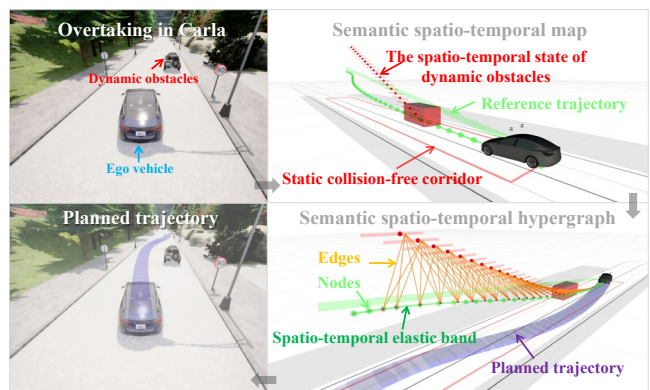

**Fig. 1.** Illustration of our trajectory generation Framework.

## I. INTRODUCTION

THE operation of autonomous vehicles in a complex urban environment presents great challenges. As one of the core modules of autonomous driving, the trajectory planning needs to provide a spatio-temporal state sequence, guiding the vehicle to pass through or avoid various obstacles to reach certain locations on time. The usual way for the trajectory planning is to establish a unified potential field [1] or occupancy grid map [2] that can project different obstacles. Then, the problem is reduced to solving non-convex path planning problems. However, these methods based on two-dimensional space faces two limitations. First, it is difficult to fully utilize the semantic information (Such as vehicles, cyclists, pedestrians and lane markings, etc.) provided by the perception module, as obstacles are represented as potential or occupancy probabilities indiscriminately. Second, it is difficult to handle dynamic obstacles because projecting trajectories of dynamic obstacles onto a two-dimensional space is a distorted representation of time-dependent constraints. Therefore, it is necessary to further consider time dimensional information besides two-dimensional space and add semantic information to obtain a semantic spatio-temporal map, which can represent complex dynamic situations.

After obtaining a semantic spatio-temporal map, we then need to search feasible and safe trajectories for autonomous vehicles. Existing methods based on state lattices [3] and spatio-temporal safety corridors [4] can produce feasible but not optimal solutions, because only a limited state space information has been utilized. Moreover, most existing methods ignore semantic information and cannot handle different types of obstacles differently.

To overcome the above problems, we propose a trajectory planning method based on multi-objective graph optimization (shown in Fig. 1). First, we construct a semantic spatio-temporal map with consideration of both static and dynamic obstacles. Second, we construct a semantic hypergraph and then reformulate the trajectory planning problem as a multi-objective graph optimization problem, which extensively expands the state space of feasible solutions. Last, we use the sparse graph optimization solver g[2]o [5] to solve the designed graph optimization problem to achieve the real-time performance. The contributions of this paper are as follows:

- A new multi-dimensional semantic spatio-temporal map that can optimize the representation of static and dynamic semantic environments is proposed.
- A spatio-temporal planner based on graph optimization that can handle multi-modal obstacles discriminatively is

Accepted final version. To Appear in IEEE Robotics and Automation Letters. ©2025 IEEE. Personal use of this material is permitted. Permission from IEEE must be obtained for all other uses. This work was supported by the National Key R&D Program of China (2023YFC3805400) and the National Natural Science Foundation of China (52172376).

[1,2]Shan He, Yalong Ma, Tao Song and Yongzhi Jiang are with School of Transportation Science and Engineering, Beihang University, Beijing, 100191, China, and also with Beijing Robint Technology Co. Ltd., Beijing, 100191, China (E-mail: shanhe@buaa.edu.cn, mayalong@buaa.edu.cn, songtaobuaa@buaa.edu.cn and yongzhijiang@buaa.edu.cn)

[1]Xinkai Wu is with School of Transportation Science and Engineering, Beihang University, Beijing, 100191, China (E-mail: xinkaiwu@buaa.edu.cn).

- proposed.
- Extensive experiments are performed to validate the proposed real-time planner, and the code will be released[1].

The remainder of this paper is organized as follows. Section II discusses the related work. Section III describes the proposed trajectory planning method. Section IV presents the details of the experiments and the results. Finally, Section V concludes the paper.

## II. RELATED WORK

### A. Spatio-temporal planning

The various trajectory planning methods proposed over the decades can be divided into two categories based on whether path (lateral) planning and velocity (longitudinal) planning are decoupled. A path-velocity decoupled planner first plans a path, and then attaches a velocity profile along the planned path (e.g., [6], [7], [8]). This cascaded process can be iterated multiple times [9] for better results. To further guarantee the solution completeness and optimality, more work focuses on coupled spatio-temporal planning that can simultaneously output paths and speeds. Ziegler *et al.* [3] made preliminary attempts by constructing a directed acyclic graph using state lattices sampled from spatio-temporal manifolds. Afterwards, improvements were made in lattice generation and search methods (e.g., [10], [11], [12]). However, these methods are difficult to achieve real-time performance due to huge computational complexity. Therefore, the sampling based Rapidly-Exploring Random Tree (RRT) and its derived methods, which have a speed advantage in high-dimensional space, are applied to spatio-temporal planning [13], [14]. Nevertheless, the accuracy of above methods is inevitably limited by the low sampling density in the state space. Furthermore, with the development of UAV planning methods in 3D space [15], Ding *et al.* [4], [17] migrated such methods to the spatio-temporal planning of unmanned vehicles. The nonlinear optimization problem with non-convex constraints can be reduced to a typical quadratic optimization problem by constructing a convex space sequence in non-convex environments and optimizing segmented Bézier curves, which greatly improves the computational efficiency while ensuring certain safety. The same method is presented in studies [18], [19], [20], [21]. However, Deolasee *et al.* [22] pointed out the significant limitation of this method is that the limited cuboidal corridor representation fails to make the most of free space for optimization, leading to the suboptimality of the solution. In addition, Gao *et al.* [23] mentioned that this type of hard constraints-based method has an inherent property that all free space in the safety corridor is treated equally, making the generated trajectories potentially very close to obstacles, as the gradient information formed by obstacles (especially dynamic obstacles) is ignored. Therefore, methods with soft constraints that use gradient information formed by obstacles, can greatly expand the space of feasible solutions, which lay the foundation for searching the optimal solution. However, so far, to the best of our knowledge, there is no spatio-temporal planning method based on soft constraint optimization.

### B. Graph optimization

Many problems in autonomous driving are sparse, such as simultaneous localization and mapping (SLAM). Based on the sparsity, these problems can be reduced from complex nonlinear optimization problems to the least squares problems, which are easily solvable [5]. For example, graph optimization is one of the commonly used methods, and its related nonlinear optimization solver g$^2$o [5] has been widely used in the SLAM field [24]. This method has been applied by Rösmann *et al.* [25] to an elastic band-based trajectory planning of low-speed robots. The flexibility and robustness exhibited by this method have made it extremely popular among the robotics community. However, due to the large amount of calculation, there is no further research so far to make it adaptable to autonomous vehicles whose speed and planned trajectory length are dozens of times higher.

## III. METHODOLOGY

### A. Overview of the framework

The framework of the proposed trajectory planning method is shown in Fig. 2. It directly receives the drivable area and trajectory prediction of semantic obstacles offered by the perception module and the trajectory point sequence (also called reference trajectory) provided by the behavior planning module, and outputs a safe, feasible trajectory that meets the vehicle kinematics and dynamics requirements to the control module.

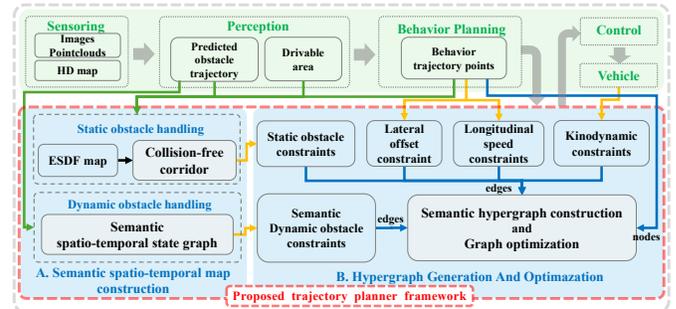

**Fig. 2.** Proposed trajectory planning framework (highlighted in red dashed box).

The proposed trajectory planning method includes two main parts. The first part is the semantic spatio-temporal map construction, including two modules: static obstacle handling and dynamic obstacle handling. The static obstacle handling first generates a Euclidean Signed Distance Field (ESDF) map that integrates the static obstacles with the drivable area of the road. Then, by iteratively querying the ESDF map around the reference trajectory points, a collision-free corridor is yielded to represent the static obstacles in a dual manner, which is the 2D part of the semantic spatio-temporal map. The dynamic obstacle handling fills the predicted trajectory of different semantic dynamic obstacles into the 3D part of the semantic spatio-temporal map. The discrete points of trajectories in a spatio-temporal coordinate system are defined as (x,y,t). The second part is to build and optimize the spatio-temporal semantic hypergraph. These two steps will be explained in the following section.

The method proposed in this paper uses the Cartesian coordinate system instead of the Frenét frame [26], as the Frenét coordinate system has the following drawbacks. Firstly, when the road centerline is an equal radius arc (especially when the radius is small), there are non-unique coordinate conversion values. Secondly, when the curvature of the road centerline is large, it is difficult to ensure the kinematic constraints of the vehicle. Last, the coordinate transformation is computationally expensive.

### B. Semantic spatio-temporal map construction

In our method, we separate static obstacles and dynamic obstacles because they have different sparsity in constraining trajectory points. Static obstacles only exert a repulsive force on trajectory points that are spatially close to them, whereas dynamic obstacles consider the additional dimension of temporal distance to adjust the trajectory. Therefore, our semantic spatio-temporal map consists of a 2D static collision-free corridor and a 3D semantic spatio-temporal state graph. Here, if the speed of the obstacle is less than 0.2 m/s, we classify it as static, otherwise it is dynamic.

**1) Static obstacle handling**

The static obstacles occupy most of the grid when overlaying the layers of non-drivable areas, static semantic obstacles, and static irregular obstacles on a grid map. If all static grids are allowed to constrain the trajectory, the computational load will increase sharply. So, the main purpose of static obstacle handling is to avoid the waste of memory space and computing resources to represent a large number of static obstacles. According to the observation, when autonomous vehicles pass by static obstacles, they only need to maintain a safe distance from the obstacles closest to both sides of their route, rather than those farther away. Therefore, we propose a two-dimensional static collision-free corridor, which expresses the existence of static obstacles in a dual (i.e., obstacle free) manner.

---

**Algorithm 1: Static Collision-free Corridor Generation**

1. **Inputs:**
   Drivable area map $M_{ob}$, Static obstacle set $O_{static}$, Reference trajectory points $\{s_0, s_1, \ldots, s_n\}_t \in SE(2)$ at time $t$, Minimum distance to obstacle $d_{obs}$, Maximum search distance $d_{max}$
2. **Initialize:**
   Boundaries $\mathcal{B}_i = \phi$ of point $s_i$, Candidate Boundaries $\mathcal{B} = \phi$, Safety Corridor $\mathcal{B}^* = \phi$
3. $M_{esdf} \leftarrow ESDFMapGeneration(M_{ob}, O_{static})$
4. **for** $s_i = s_0, s_1, \ldots, s_n$ **do** // Traverse every point
5. $\quad \mathcal{B}_i \leftarrow BoundarySearch(M_{esdf}, s_i, d_{obs}, d_{max})$
6. $\quad \mathcal{B} \leftarrow \mathcal{B} \cup \mathcal{B}_i$
7. **end for**
8. $\mathcal{B}^* \leftarrow BoundaryAssociation(\mathcal{B})$
9. **return** $\mathcal{B}^*$

---

The algorithm for generating collision-free corridor is shown in Algorithm 1. It includes the following sub modules:

*ESDFMapGeneration*: This function projects the drivable areas of the road and static obstacles onto a grid map, and further calculates the Euclidean Signed Distance. Ultimately, the map will contain interval distance information from each obstacle free grid to the nearest obstacle grid.

*BoundarySearch*: This function queries the ESDF map from right to left at a certain step size in the direction perpendicular to orientation of each trajectory point, and records grid points with a distance of $d_{obs}$ (mentioned in the Line 1 of Algorithm 1) from the obstacle (see Fig. 3).

*BoundaryAssociation*: This function pairs the left and right boundary points of each trajectory point in the vertical direction. When a trajectory point has multiple boundary pairs, it will choose the boundary pair with a high overlap rate with the boundary pair of the previous trajectory point. Finally, when connecting the left and right boundaries of all trajectory points, a closed safety corridor (see Fig. 3) is generated. It is worth mentioning that the collision-free corridor connected by discrete points can be made continuous through interpolation methods.

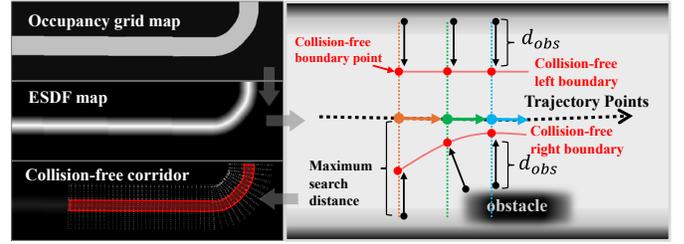

**Fig. 3.** Illustration of 2D collision-free corridor generation. The area enclosed by the solid red line in the lower left picture is a collision-free corridor obtained on the ESDF map. The image on the right shows the details during the generation of collision-free corridor. The red solid line is connected by red dots at the minimum distance from the obstacle $d_{obs}$.

To achieve collision free trajectory planning, we must establish constraints in linear equations for vehicles to be inside the corridor and not to collide with obstacles. For linearity, we use three circles with the same radius $r$ to approximate the shape of the ego vehicle [27] as shown in Fig. 4 (a), which is reliable and easy to implement. We formulate the linear constraints where those three circles on each trajectory point is collision-free.

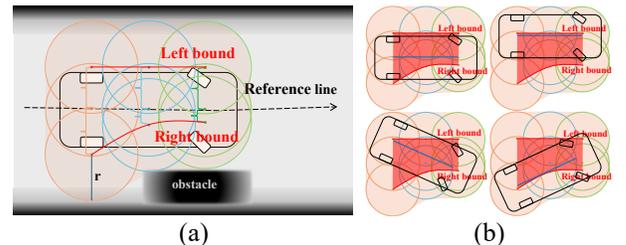

(a)      (b)

**Fig. 4.** Ego Vehicle collision model. (a) shows the ego vehicle collision model represented by three circles with radius $r$, where the centers of the three circles need to be within the collision-free corridor. (b) shows that as long as the blue line formed by the centers of those three circles is within the red area, the vehicle will not collide with the obstacle.

**2) Dynamic Obstacle Handling**

Typically, dynamic obstacles include moving vehicles, cyclists, and pedestrians. We simplify their predicted spatio-temporal poses through separate collision models and place them in a spatio-temporal domain composed of $(x, y, t)$ (see Fig. 5), making it easier to search when constructing associations with the trajectory points of ego vehicle. It should be noted that the interaction effects of changes in the trajectory of ego vehicles on dynamic obstacles have been taken into account in behavior planning. In this work, we assume that the predicted trajectory of dynamic obstacles is fixed, and only the trajectory of ego vehicle will be affected.

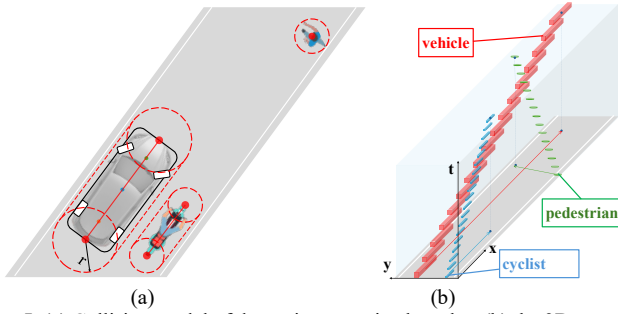

**Fig. 5.** (a) Collision model of dynamic semantic obstacles. (b) the 3D semantic spatio-temporal state graph.

The collision model is pill shaped (see Fig. 5), consisting of a line segment and a circle whose center can move on the line segment. The two endpoints of the line segment are the center points of the front and rear of the detected semantic bounding box, and the radius of the circle is equal to half the width of the bounding box. This allows for the preservation of the perceived length and width of obstacles while being simpler than a rectangular collision detection model. Note the ego vehicle needs to maintain a distance greater than the radius of the circle from the line segment when planning its trajectory.

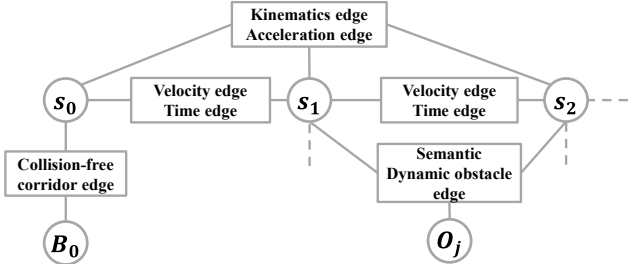

**Fig. 6.** Hyper-Graph structures. The nodes (circles) and edges (rectangles) respectively represent the variables and sub-cost functions (soft constraint equations) in graph optimization. $s_0$, $s_1$, $s_2$ represent three consecutive nodes to be optimized (in this context, they refer to trajectory points). $B_0$ is the boundary of the collision-free corridor corresponding to $s_0$. $O_j$ represents the $j$-th dynamic obstacle.

## C. Hypergraph generation and graph optimization

Here, we construct a hypergraph of sparse connections between obstacles and ego vehicle trajectory points and between ego vehicle trajectory points (see Fig. 6), and then use the g²o solver to optimize the graph to find the solution. We also call our method a spatio-temporal elastic band planner (STEB planner) as the generation of feasible trajectories during graph optimization is like deforming a spatio-temporal elastic band under various constraints.

The first task is to determine the hypergraph nodes. The classic trajectory planning [25] uses robot state sequence consisting of $\{x, y, \theta\} \in SE(2)$ in two-dimensional Special Euclidean Group as nodes. To adapt to dynamic environments, the variables are expanded to $\{x, y, \theta, t\}$. Aiming to decrease computational complexity, we obtain $\theta$ through the direction formed by two adjacent state points by (1), which can reduce the number of optimization variables and eliminate additional non holonomic motion constraints. Therefore, the final optimization variable was set to $s_k = \{x_k, y_k, t_k\}$, also called a STEB node, initialized from the reference trajectory. In addition, we define the pose point of reference trajectory $S' = \{s'_k\}_{k=1,\cdots,m}, m \in \mathbb{N}$ as $s'_k = \{x'_k, y'_k, \theta'_k, t'_k\}$.

$$\theta_k = arctan\left(\frac{y_{k+1} - y_k}{x_{k+1} - x_k}\right) \quad (1)$$

The key idea of our proposed method is to adjust and optimize STEB nodes through multi-objective optimization methods by a weighted sum model as follows:

$$S^* = \underset{S}{\mathrm{argmin}} \sum_i \alpha_i f_i(S) \quad (2)$$

where $S^*$ denotes the optimized STEB nodes, $S = \{s_k\}_{k=1,\cdots,n}$, $n \in \mathbb{N}$, $f_i(S)$ is the component objective functions, and $\alpha_i$ is its weight. It is worth noting that the cost formed by the $f_i(S)$ terms in multi-objective optimization is considered a soft constraint. For the sake of convenience, the constraint terms mentioned later in the text refer to the sub-objective functions here.

The second step is to construct the edges of the hypergraph. There are two types of constraints for multi-objective optimization functions: continuous constraints and piecewise continuous constraints. The error functions for discontinuous constraints can be found in (3), (4), (5), and the specific constraint edges we constructed are as follows:

$$E_{inter}(x_k, x_{min}, x_{max}) = \begin{cases} 0 & x_{min} \le x_k \le x_{max} \\ x_k - x_{max} & x_k > x_{max} \\ x_{min} - x_k & x_k < x_{min} \end{cases} \quad (3)$$

$$E_{more}(x_k, x_{min}) = \begin{cases} 0 & x \ge x_{min} \\ x_{min} - x & x < x_{min} \end{cases} \quad (4)$$

$$E_{less}(x_k, x_{max}) = \begin{cases} 0 & x \le x_{max} \\ x_k - x_{max} & x > x_{max} \end{cases} \quad (5)$$

Where $x_{min}$ and $x_{max}$ are parameters that predefine the minimum and maximum limits.

**C1:** Shortest path distance constraint:

$$f_1(S) = \sum_{k=1}^{n-1} \|(x_{k+1} - x_k, y_{k+1} - y_k)\| \quad (6)$$

**C2:** Kinematic (minimum turning radius $r_{min}$) constraint:

$$f_2(S) = \sum_{k=1}^{n-2} E_{more}(r_k, r_{min}) \quad (7)$$

$$r_k = \frac{\|(x_{k+1} - x_k, y_{k+1} - y_k)\|}{2 \times \sin((\vartheta_{k+1} - \vartheta_k) \times 0.5)} \quad (8)$$

**C3:** Time constraints: including efficiency (less time spent) and to avoid a decrease in trajectory points over time:

$$f_3(S) = \sum_{k=1}^{n-1} (t_{k+1} - t_k) \quad (9)$$

$$f_4(S) = \sum_{k=1}^{n-1} E_{more}(t_{k+1} - t_k, 0) \tag{10}$$

**C4:** Dynamic constraints: To meet the maximum acceleration and deceleration constraints while improving vehicle comfort, including velocity constraints (11-12), acceleration constraints (13-14), and jerk constraints (15):

$$f_5(S) = \sum_{k=1}^{n-1} E_{inter}(v_k, v_{min}, v_{max}) \tag{11}$$

$$f_6(S) = \sum_{k=1}^{n-1} E_{inter}(v_k^{ang}, v_{min}^{ang}, v_{max}^{ang}) \tag{12}$$

$$f_7(S) = \sum_{k=1}^{n-2} E_{inter}(a_k, a_{min}, a_{max}) \tag{13}$$

$$f_8(S) = \sum_{k=1}^{n-2} E_{inter}(a_k^{ang}, a_{min}^{ang}, a_{max}^{ang}) \tag{14}$$

$$f_9(S) = \sum_{k=1}^{n-3} jerk_k, \quad jerk_k = \frac{|a_{k+1} - a_k|}{|t_{k+1} - t_k|} \tag{15}$$

where:

$$v_k = \frac{\|(x_{k+1} - x_k, y_{k+1} - y_k)\|}{|t_{k+1} - t_k|} \tag{16}$$

$$v_k^{ang} = \frac{|\theta_{k+1} - \theta_k|}{|t_{k+1} - t_k|} \tag{17}$$

$$a_k = \frac{v_{k+1} - v_k}{|t_{k+1} - t_k|} \tag{18}$$

$$a_k^{ang} = \frac{v_{k+1}^{ang} - v_k^{ang}}{|t_{k+1} - t_k|} \tag{19}$$

**C5:** Collision-free corridor boundary constraints derived from static obstacles:

$$f_{10}(S) = \sum_{k=1}^{n} E_{inter}(d_k, d_{min}, d_{max}) \tag{20}$$

where: $d_k$ is the lateral offset of the centers of the three circles of the ego vehicle's collision model relative to their own nearest reference point $s'_{nea} = \{x'_{nea}, y'_{nea}, \theta'_{nea}, t'_{nea}\}$:

$$d_k = \frac{(x_k - x'_{nea}, y_k - y'_{nea}) \times (\cos(\theta'_{nea}), \sin(\theta'_{nea}))}{\|(\cos(\theta'_{nea}), \sin(\theta'_{nea}))\|} \tag{21}$$

**C6:** Dynamic obstacles constraints, including:

1). Maintaining the minimum spatio-temporal distance $D_{min}$ results in achieve the effect of exchanging space for time, which is a more active form of interaction with dynamic objects $O^j$ like vehicles:

$$f_{11}(S) = \sum_{k=1}^{n} E_{more}(d_{k,j}, D_{min}) \tag{22}$$

$$d_{k,j} = \|(x_k - O_{t,x}^j, y_k - O_{t,y}^j, t_k - O_t^j)\| \tag{23}$$

where: $(O_{t,x}^j, O_{t,y}^j, O_t^j)$ is the coordinate of the obstacle $O^j$ in the spatio-temporal coordinate system at time $t$.

2). Maintaining the minimum temporal distance $T_{min}$ only adjusts the speed, which is a relatively conservative form of interaction with dynamic objects $O^j$ like pedestrians:

$$f_{12}(S) = \sum_{k=1}^{n} E_{more}(d_{k,j}, T_{min}) \tag{24}$$

$$d_{k,j} = |t_k - O_t^j| \tag{25}$$

C6 constraint edges is generated when the spatial distance and temporal distance between the dynamic obstacles and the STEB nodes are less than $\beta_d D_{min}$ and $\beta_t T_{min}$. $\beta_d$ and $\beta_t$ are proportional coefficients. Specifically, as the obstacle avoidance distance increases, there is a corresponding increase in the quantity of edges generated in the graph structure.

**C7:** Distance constraint from reference trajectory:

$$f_{13}(S) = \sum_{k=1}^{n} \|(x_k - x'_k, y_k - y'_k)\| \tag{26}$$

$$f_{14}(S) = \sum_{k=1}^{n} |(t_k - t'_k)| \tag{27}$$

**C8:** Current state constraints, including start speed constraints and start direction constraints composed of the current vehicle state:

$$f_{15}(S) = \sum_{k=1}^{l} |v_k - v_{current}| \tag{28}$$

$$f_{16}(S) = \sum_{k=1}^{l} |\theta_k - \theta_{current}| \tag{29}$$

where: $l$ represents the number of nodes to which the constraint is to be applied.

**C9:** Goal point state constraints, including arrival time difference constraints, orientation constraints, and distance constraints formed by the goal point:

$$f_{17}(S) = \|(x_n - x'_m, y_n - y'_m)\| \tag{30}$$
$$f_{18}(S) = |(\theta_n - \theta'_m)| \tag{31}$$
$$f_{19}(S) = |(t_n - t'_m)| \tag{32}$$

Most of the above sub-objective functions are local and only related to a small number of consecutive STEB nodes. This local structure produces a sparse system matrix, allowing solutions to be found quickly and efficiently.

## IV. EXPERIMENTAL RESULTS

### A. Implementation

TABLE I
PARAMETER SETTINGS IN THE EXPERIMENT

| $\alpha_1$ | $\alpha_2$ | $\alpha_3$ | $\alpha_4$ | $\alpha_5$ | $\alpha_6$ | $\alpha_7$ | $\alpha_8$ |
|---|---|---|---|---|---|---|---|
| 10 | 1.0 | 1.0 | 100 | 10 | 100 | 100 | 50 |
| $\alpha_9$ | $\alpha_{10}$ | $\alpha_{12}$ | $\alpha_{13}$ | $\alpha_{14}$ | $\alpha_{15}$ | $\alpha_{16}$ | $\alpha_{17}$ |
| 100 | 200 | 200 | 0.1 | 0.1 | 0.1 | 0.1 | 1.0 |
| $\alpha_{18}$ | $\alpha_{19}$ | $r_{min}$ | $D_{min}$ | $T_{min}$ | $\beta_d$ | $\beta_t$ | $a_{max}^{ang}$ |
| 10 | 30 | 5.0 | 3.0 | 4.0 | 1.5 | 1.5 | 0.5 |

We validate our method in the CARLA [16] simulation with a frequency of 20Hz. To demonstrate the robustness of the proposed algorithm for practical applications, we use the current position and bounding box of the obstacles output by the simulation, and the obstacle tracking is implemented by the Kalman filter, while trajectory prediction is generated by a constant velocity model based on High Definition (HD) map. To ensure the objectivity and fairness of experiments and analyses, our behavioral planning module, implemented as a simple finite state machine, only outputs reference trajectories with uniform longitudinal speed along with the routing path without making decisions that may interfere with the experiment. Additionally, the spatial resolution of the reference trajectory is set to 0.5 meters.

Adjusting the weight of each sub objective in multi-objective optimization is an experiential task. Table I provides the weight values of each component objective in our experiments. All the experiments are conducted on a desktop computer with an Intel 11900K CPU (3.5 GHz).

*B. Qualitative Results*

To demonstrate that the proposed method can cope with various common static and dynamic scenarios, the following representative experimental cases are conducted.

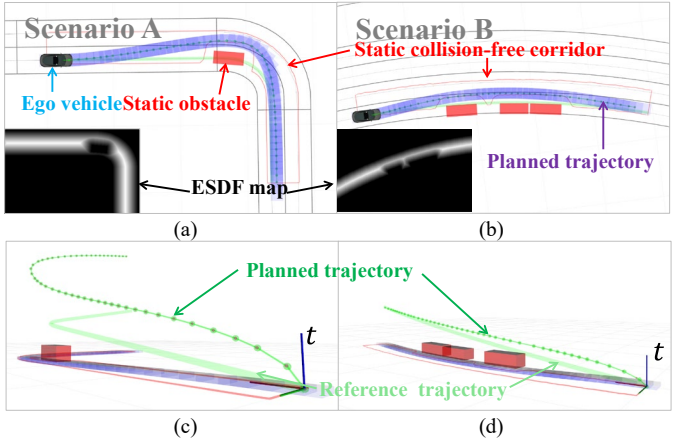

**Fig. 7.** Scenarios of bypassing stationary obstacles. (a) Scenario A: a car stops at a corner of the road; (b) Scenario B: multiple vehicles park on the roadside. (c) and (d) are the 3D views of trajectory planning results for static obstacles.

**1) Static obstacle avoidance (Scenario A&B)**

As shown in Fig. 7, we conducted experiments in two scenarios. The ego car needs to deal with the stationary obstacles in front (one or more parked cars in the red box). It can be seen from Figs. 7(a) and (b) that the continuous smooth trajectories planned by our method can maintain an appropriate distance from the obstacles. Figs. 7(c) and (d) show that our method considers the current state of the vehicle in speed planning and reduces speed in curved sections.

**2) Unprotected left turn (Scenario C)**

As shown in Figs. 8(a) and (b), the ego car in the blue box is ready to turn left following the blue arrow, but the car needs to negotiate with the consecutive obstacle cars in the red box traveling along the red arrow. This scenario indicates the interactive ability of the proposed method for the ego vehicle with vehicles from different directions. Experimental results show that our method can quickly find a feasible and safe trajectory in continuous opposing traffic flows.

**3) Merging into dense traffic (Scenario D)**

This case demonstrates the ability of our proposed method to interact with relatively high-speed traffic in the same traveling direction. According to Figs. 8(c) and (d), we can see that the ego car on the ramp plans a trajectory that can merge into the main road. After merging into the traffic flow, the ego car begins to follow the preceding vehicle by adjusting its speed.

**4) Overtaking on an urban roadway (Scenario E)**

In this scenario, ego vehicle needs to borrow the opposite lane on a two-way two-lane to overtake the slow-moving vehicle ahead. To ensure safety, it is necessary to quickly overtake the preceding vehicle and leave the opposite lane to return to the lane of the vehicle. As can be seen from Figs. 8(e) and (f), our method can complete the overtaking task quickly while ensuring a safe distance between vehicles.

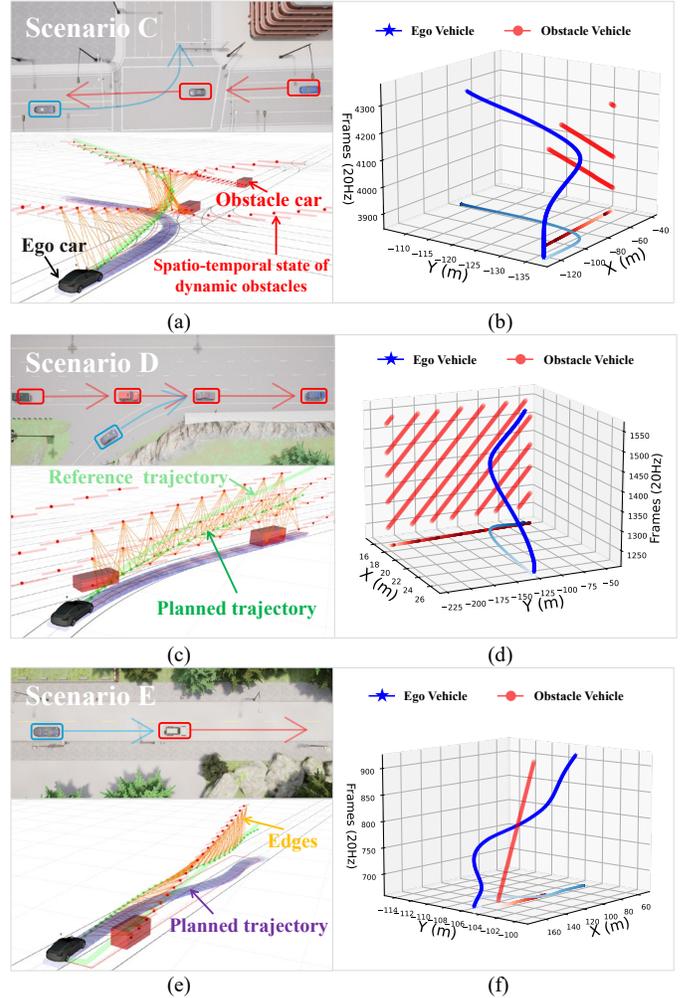

**Fig. 8.** Scenarios with dynamic obstacles. (a, c, e) are three experimental scenarios in Carla and the screenshots of the real-time trajectory planning of the proposed method in spatio-temporal dimensions. The red cube, red point, and red line segment represent the current posture, trajectory prediction results and corresponding collision model of the obstacle vehicle in the spatio-temporal dimension, respectively. The two ends of the orange line segment connect the space-time points of the obstacle prediction trajectory and the ego car's planned trajectory, indicating that the obstacle space-time point has an influence on the self-vehicle trajectory point (that is, the edge of the hypergraph is generated). (b, d, f) are visualizations of the continuous changes in the positions of the ego and obstacle cars. The gradient blue thin lines and red thin lines are the trajectory projections (i.e., paths) of the ego and obstacle cars, respectively.

*C. Quantitative Results*

This section will quantitatively verify that the proposed method can enable different semantic elements to impose different constraints on the planned trajectory, thereby meeting various requirements, such as safety. We will compare it with a similar method SSC proposed by Ding *et al.* [4], which constructs hard constraints for quadratic optimization problems using spatio-temporal corridors in the Frenét frame. We design scenarios $F_1$ and $F_2$ as shown in Fig. 9. For a fair comparison, the following configuration is made:

- **Same experimental scenario:** At the configuration depicted in Fig. 9, the dynamic obstacle and the ego vehicle are traveling at a velocity of $3\ m/s$. Maintaining this velocity would result in a collision at the identified conflict point. To evaluate different scenarios, we modified the longitudinal velocity of the reference trajectory to perform two different operations: **Proceed**: The ego vehicle accelerates to traverse the conflict zone prior to the dynamic obstacle's arrival. In this scenario, the longitudinal velocity is set to $5\ m/s$. **Yield**: The ego vehicle decelerates to cede right-of-way to the dynamic obstacle. Here, the longitudinal velocity is set to $2\ m/s$.
- **Same information input:** Two methods have the same parameter configuration and information input set, including the same behavior trajectory points input, the same road speed limit ($10\ m/s$), and the same maximum acceleration ($2\ m/s^2$) and deceleration ($3\ m/s^2$), etc.

We selected two types of traffic participation, vehicles and pedestrians, with different attributes (the advantaged group using dynamic obstacle constraint (22) and the vulnerable group using (24) to conduct experiments. To evaluate and compare the planned trajectories, we employ three key metrics:

**Safety**: Post Encroachment Time (PET) [28]. This metric measures the temporal interval between two vehicles consecutively entering a conflict zone. A larger interval indicates a higher degree of safety.

**Efficiency**: Mean velocity of the planned trajectory.

**Comfort**: Maximum absolute jerk of the planned trajectory.

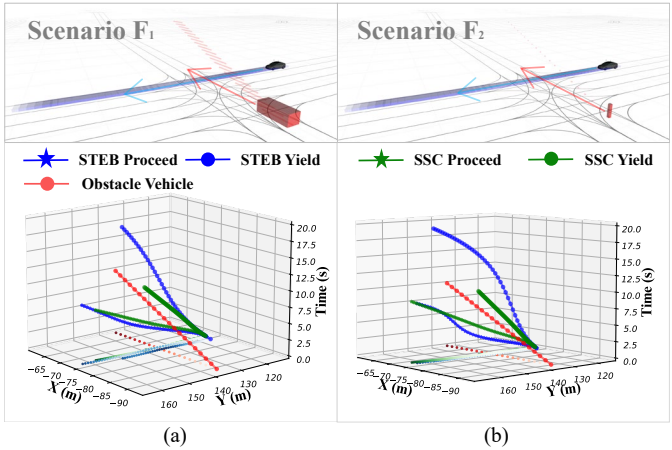

**Fig. 9.** Illustrations of different semantic obstacle negotiation scenarios and trajectory planning results under them.

### 1) Comparison of planned trajectories

Experiments were conducted on the proposed STEB and SSC with the two aforementioned dynamic obstacles under different behavior planning directives (proceed and yield). The resultant planned trajectories are illustrated in Fig. 9, with corresponding metric analyses presented in Table II. The data reveal that STEB has a larger PET compared to SSC, with notably larger PET values observed during pedestrian interactions relative to vehicle interactions. This demonstrates that the trajectories planned by our method offer better safety profile compared to SSC. Furthermore, it highlights STEB's capacity to differentially respond to diverse semantic objects based on the imposed trajectory constraints.

TABLE II
TRAJECTORY EVALUATION OF SCENARIO F

|  |  |  | PET (s) | Avg. Vel. (m/s) | Max. Jerk (m/s³) |
|---|---|---|---|---|---|
| **STEB** | Ve. | Pro. | **2.97** | **5.65** | 3.39 |
|  |  | Yie. | **5.88** | 2.20 | 0.54 |
|  | Pe. | Pro. | **3.51** | **6.56** | 10.23 |
|  |  | Yie. | **7.90** | **2.87** | 0.86 |
| **SSC** | Ve. | Pro. | 2.19 | 5.61 | **2.32** |
|  |  | Yie. | 2.14 | 2.48 | **0.35** |
|  | Pe. | Pro. | 2.11 | 5.59 | **2.32** |
|  |  | Yie. | 2.25 | 2.48 | **0.35** |

Meanwhile, the average velocities of the STEB trajectory are not significantly deviate from those of SSC (slightly higher in some cases), indicating that our method can maintain high efficiency while providing better safety.

It is important to acknowledge that SSC, based on polynomial Bézier curve optimization, demonstrates superior comfort metrics. Notably, when the ego vehicle is required to traverse the conflict zone prior to pedestrian passage, STEB exhibits large jerk values. This phenomenon can be attributed to STEB's parameter settings, which necessitates rapid acceleration of the ego vehicle to ensure a safer separation distance from pedestrians, followed by deceleration post-conflict zone to adhere to speed limit constraints. In this scenario, reducing the weights of obstacle constraints C6 or increasing the weights of dynamic constraints C4 can result in a more comfortable trajectory, but it may have the side effect of lowering the PET value. These adjustments can be customized according to user needs.

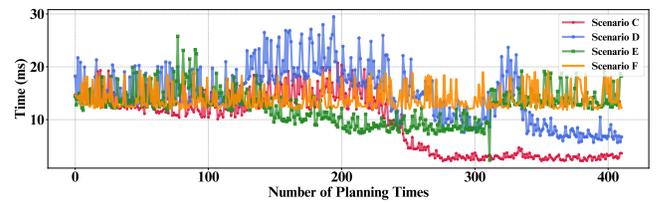

**Fig. 10.** STEB time consumption (ms) in different scenarios.

### 2) Real-time Performance

Fig. 10 and Table III show the computational time of our proposed method in each scenario. The average, maximum, and minimum single run time of our algorithm in all experiments are 12.63ms, 29.48ms, and 2.9ms, respectively. The maximum value occurs when merging into dense traffic flow from a curved ramp. In addition, in the experiment of scenario F, our average and maximum time consumption are better than those of SSC. Overall, this shows that our method has real-time computing performance.

TABLE III
TIME CONSUMPTION OF PLANNERS IN DIFFERENT SCENARIOS

| Time (ms) | STEB | | | | SSC |
|---|---|---|---|---|---|
| | Sce. C | Sce. D | Sce. E | Sce. F | Sce. F |
| Avg. | 9.48 | 14.67 | 12.44 | 13.95 | 19.03 |
| Min. | **2.18** | 5.67 | 2.31 | 11.82 | 9.39 |
| Max. | 20.77 | **29.48** | 25.78 | 19.37 | 37.90 |

## V. CONCLUSION AND FUTURE WORK

In this paper, we propose a trajectory planning method for autonomous vehicles based on multi-objective graph optimization. This method innovatively uses two-dimensional safety corridors and semantic spatio-temporal state graph to respectively represent the spatial constraints of static obstacles and spatio-temporal constraints of dynamic obstacles on trajectory planning. Then, by constructing a semantic spatio-temporal hypergraph, we transform trajectory planning problem into a multi-objective optimization problem and solve it using graph optimization theory. At the end, extensive experiments have been conducted to demonstrate that our method can utilize multi-modal results output from perception to cope with complex urban scenarios and achieves real-time computational performance.

In the future, we will design a matching behavior planner, which can provide better initial values for the proposed method and adjust its parameters as the scene changes to achieve better performance. Additionally, more types of constraints (such as social relationships) can be incorporated into the edges of graph optimization to further enhance trajectory planning.